\documentclass[conference]{IEEEtran}
\usepackage{times}

\usepackage[numbers]{natbib}
\usepackage{multicol}
\usepackage[bookmarks=true]{hyperref}
\usepackage{graphicx, amsmath, amssymb}
\usepackage{xcolor}
\usepackage{cleveref}
\usepackage{wrapfig}
\usepackage{soul}

\begin{document}

\title{Logically Constrained Robotics Transformers for Enhanced Perception-Action Planning}


\author{
\IEEEauthorblockN{Parv Kapoor\IEEEauthorrefmark{1}}
\IEEEauthorblockA{\textit{Carnegie Mellon University} \\
Pittsburgh, PA USA \\
parvk@andrew.cmu.edu}
\and
\IEEEauthorblockN{Sai Vemprala}
\IEEEauthorblockA{\textit{Scaled Foundations} \\
Seattle, WA USA \\
sai@scaledfoundations.ai}
\and
\IEEEauthorblockN{Ashish Kapoor}
\IEEEauthorblockA{\textit{Scaled Foundations} \\
Seattle, WA USA \\
ashish@scaledfoundations.ai}
}



%

\maketitle
\begingroup
\renewcommand\thefootnote{\IEEEauthorrefmark{1}}
\footnotetext{Work done while author was an intern at Scaled Foundations}
\endgroup
\begin{abstract}
With the advent of large foundation model based planning, there is a dire need to ensure their output aligns with the stakeholder's intent. When these models are deployed in the real world, the need for alignment is magnified due to the potential cost to life and infrastructure due to unexpected faliures. Temporal Logic specifications have long provided a way to constrain system behaviors and are a natural fit for these use cases. In this work, we propose a novel approach to factor in signal temporal logic specifications while using autoregressive transformer models for trajectory planning. We also provide a trajectory dataset for pretraining and evaluating foundation models.  Our proposed technique acheives 74.3 \% higher specification satisfaction over the baselines.
\end{abstract}

\IEEEpeerreviewmaketitle

\section{Introduction}



Most autonomous robots deployed in the real world need to operate in a safe and reliable manner. This problem is exacerbated for safety critical applications such as autonomous vehicles where failure to respect safety constraints can lead to catastrophic consequences. A key issue with ensuring safety constraints can be attributed to imprecise specification of desired behaviors. Existing methods of encoding behaviors through objectives, such as cost functions or reward functions, can be exploited by underlying algorithms in a suboptimal manner. \cite{clark2016faulty} This allows them to achieve high scores without fully meeting the intended requirements. This problem is only worsened by the recent advent of using Natural Language (NL) to communicate instructions to robots. Due to NL's inherent ambiguity, it is unclear how to use it for encoding precise behavior or constraints in safety critical applications. \cite{vemprala2024chatgpt, tellex2020robots}

An alternative is to use specifications written in Temporal logics (TL) that have rich and precise semantics. Additionally, TL specifications provide a tractable way to check if the system achieves the desired behavior. In line with this, there has been significant recent interest in specifying temporal and logical constraints on system behaviors through a continuous-time real-valued TL called  \emph{signal temporal logic (STL)} \cite{Maler2004MonitoringTP}. STL specifications can be defined over state action trajectories of robots. Additionally, STL is equipped with a score of satisfaction/violation called \textit{Robustness} that can be used as feedback for generating desired behavior. There exist a large body of work that uses STL specifications to generate behavior through Reinforcement Learning \cite{aksaray2016qlearning, kapoor2020model}, Mixed Integer Convex Programming (\cite{kurtz2022mixed,kapoor2024safe}), Monte Carlo Tree Search \cite{10160953}, and gradient based techniques (\cite{leungbackpropagation}). 
However, despite the recent widespread use of foundation models for behavior planning, their application in generating behaviors that satisfy STL specifications has not yet been explored.
\begin{figure}[t]
    \centering
    \includegraphics[width=0.5\textwidth]{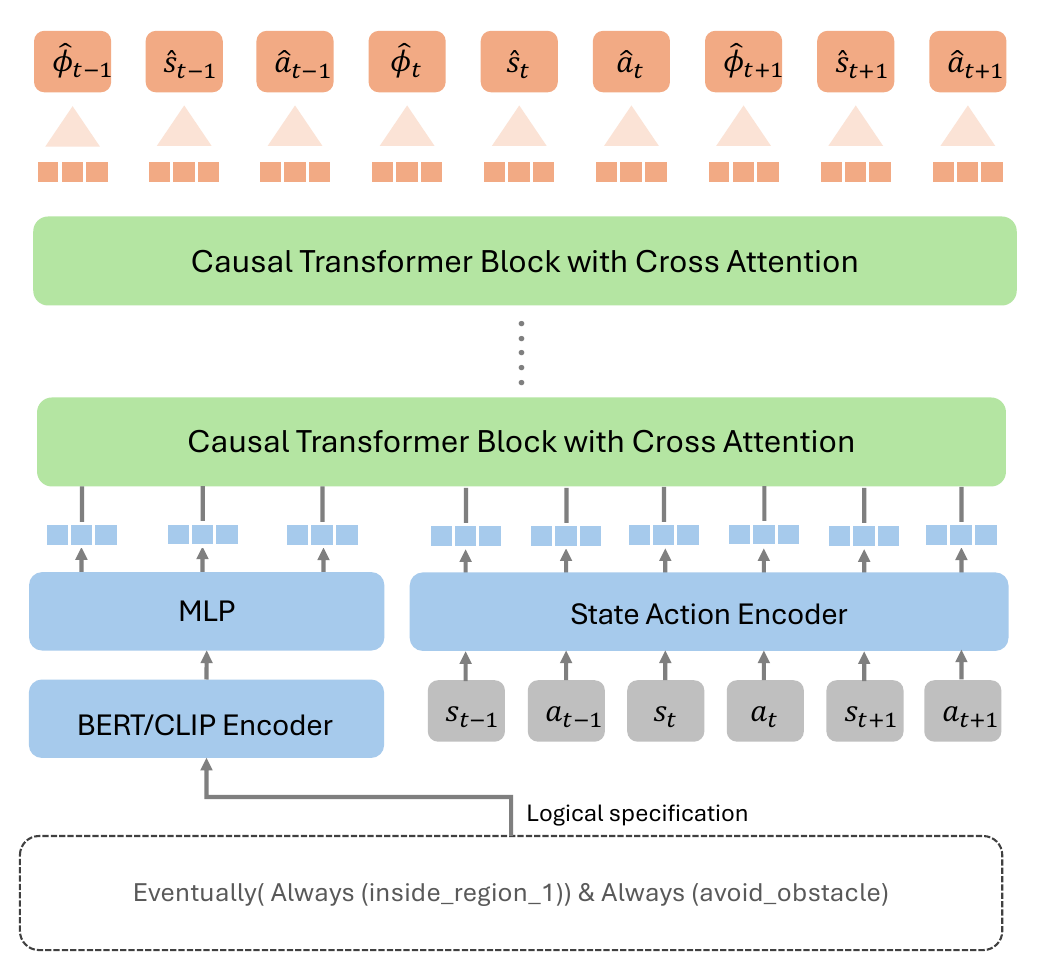}
    \caption{The PASTEL Architecture utilizes both causal and cross attention mechanisms to autoregressively predict Signal Temporal Logic (STL) specification satisfying trajectories conditioned on state, action, and specification embeddings. }
    \label{fig:wide_figure}
\end{figure}
This paper explores the integration of STL specifications with data-driven foundation model based trajectory planning. Specifically, we leverage the Perception-Action Causal Transformer (PACT) \cite{bonatti2023pact} that uses an autoregressive transformer model to learn robot trajectories in an unconstrained fashion during pretraining. Then, the pretrained PACT model is used to predict trajectories during runtime, inherently ensuring safety due to the distribution of safe trajectory data. In this work, We augment PACT with STL specifications to achieve constrained trajectory planning. By leveraging the strengths of large pretrained models and formal verification, we aim to bridge the gap between data driven planning and concrete specification satisfaction for enhanced safety and liveness.

Our approach involves explicitly factoring in STL specifications into the autoregressive trajectory planning process. Specifically, we employ state of the art language tokenizers such as Contrastive Language Image Pretraining (CLIP) to generate embeddings for the STL specifications. Since, PACT excels at integrating multi modal inputs in the embedding space for trajectory prediction, we repeat the specification embedding and append it to the state action embeddings at each time step. Second, Inspired by Vision Language Action (VLA) models \cite{ma2024survey}, we extend PACT with a cross attention mechanism where specification embeddings are used as queries and state action embeddings are used as keys and values. Then, we define a specification relevance loss based on a similarity metric over the cross attention outputs and the text embedding to force the model to "attend" to the specification tokens while making predictions. 

To pretrain and evaluate our model, we create a dataset of specifications and state action trajectories using STLPy \cite{kurtz2022mixed} for a 2D planar environment. These specifications are designed from common motion planning patterns defined over atomic propositions encoding occupancy of different regions in the state space.

In summary, this paper presents a novel framework that combines the precision of STL specifications with power of large pretrained models for robotic planning. To the best of the authors’
knowledge, this is the first approach integrating STL specifications with causal transformers for trajectory planning.
\\
The key contributions of the paper are:
\begin{itemize}
    \item A novel specification tokenization and cross attention based approach for requirement satisfaction
    \item A state, action, specification trajectory dataset for pretraining foundation models
    \item An evaluation of the proposed approach over a benchmark of motion planning tasks.
\end{itemize}


\section{Preliminaries and Problem Statement}

\subsection{Signal Temporal Logic}
STL is a logic specification language used to define properties of continuous time real valued signals \cite{Maler2004MonitoringTP}.
A signal $\mathbf{s}$ is a function $\mathbf{s}:T \to \mathbb{R}^n$ that maps a time domain $T \subseteq \mathbb{R}_{\geq 0}$ to a real valued vector. Then, an STL formula is defined as:
$$\phi := \mu ~|~ \neg \phi ~|~ \phi \land \psi ~|~ \phi \lor \psi ~|~ \phi ~\mathcal{U}_{[a,b]}~\psi$$
where $\mu$ is a predicate on the signal $\mathbf{s}$ at time $t$ in the form of $\mu \equiv \mu(\mathbf{s}(t)) > 0$ and $[a, b]$ is the time interval. The \emph{until} operator $\mathcal{U}$ defines that $\phi$ must be true until $\psi$ becomes true within a time interval $[a, b]$. 

Given a signal $s_{t}$ representing a signal starting at time t, the Boolean semantics of satisfaction of $s_t \models \phi$ are defined inductively as follows:
\begin{align*}
    s_t\models\mu  &  \iff   \mu(s(t))>0 \\
     s_t \models \lnot \varphi & \iff  \lnot (s_t \models \varphi ) \\
    s_t \models \varphi_1 \land \varphi_2 & \iff  (s_t \models \varphi_1) \land (s_t \models \varphi_2) \\
  s_t \models \text{F}_{[a,b]}(\varphi) & \iff \exists t' \in [t+a, t+b] \text{ s.t. }  s_{t'} \models \varphi \\
     s_{t} \models \text{G}_{[a,b]}(\varphi) &\iff  \forall t' \in [t+a, t+b] \text{ s.t. }  s_{t'} \models \varphi 
\end{align*}

\subsection{Problem Statement}
Consider a dynamical system with states \( x_t \in \mathbb{R}^n \) and actions \( a_t \in \mathbb{R}^m \) at discrete time steps \( t \). The objective is to predict a sequence of state-action pairs \( \{(x_t, a_t)\}_{t=0}^T \) over a finite horizon \( T \) such that the predicted trajectory satisfies a given STL specification \( \varphi \).

Our system dynamics are defined by $x_{t+1} = f(x_t, a_t)$ where \( f: \mathbb{R}^n \times \mathbb{R}^m \rightarrow \mathbb{R}^n \) maps a state action pair $(x_t \in \mathbb{R}^n , a_t \in \mathbb{R}^m)$ at a given timestep $t$ to the next state $x_{t+1} \in \mathbb{R}^n$.

For generating trajectories, we utilize a causal transformer model to autoregressively predict the next state-action pair based on the past sequence:
\begin{equation}
   \{(x_{t+k}, a_{t+k})\}_{k=1}^{T-t} = \text{Transformer}(\{(x_i, a_i)\}_{i=0}^{t+k-1}, \theta)
\end{equation}
where \( \theta \) represents the parameters of the transformer model.

Then, An STL specification \( \varphi \) is defined over the state and action variables encoding the safety and liveness requirement. A trajectory \( \{(x_t, a_t)\}_{t=0}^T \) satisfies \( \varphi \) if:
\begin{equation}
    \{(x_t, a_t)\}_{t=0}^T \models \varphi
\end{equation}

The goal is to find the parameters \( \theta \) of the causal transformer model such that the predicted trajectory \( \{(x_t, a_t)\}_{t=0}^T \) maximizes the likelihood of satisfying the STL specification \( \varphi \).

Mathematically, this can be expressed as:

\begin{equation}
\begin{aligned}
\theta^* &= \arg\max_{\theta} \Pr( {(x_t, a_t)}_{t=0}^T \models \varphi \mid \theta ) \\
\textrm{s.t.} \quad & x_{t+1} = f(x_t, a_t) \quad \forall t \in {0, 1, \ldots, T-1} \\
& x_0 = x_{\text{init}}, \quad a_0 = a_{\text{init}} \\
\end{aligned}
\end{equation} \\

\section{Methodology}

\subsection{Approach}
The overall approach is described in Figure 1. We call our model Perception Action Signal TEmporaL Transformer (PASTEL). PASTEL takes as input the state, action and specification embeddings and autoregressively predicts the future sequence of states and actions conditioned on the embeddings such that the state trajectory signal satisfies the user specified STL specification. The length of the state-action trajectory is fixed as the horizon of the original specification keeping in line with the original STL semantics. 
\begin{table*}[!ht]
    \centering

\begin{tabular}{|p{0.5cm}|p{7cm}|c|}
\hline
 &  STL Specification & Pattern \\
\hline
 $\phi_1$ & $F_{[0,10]} (R_1) \lor F_{[10,20]} (R_2)\land F_{[20,30]} (R_3) \land G_{[0,30]} (\neg O_1)$ & MRC+A \\
\hline
 $\phi_2$ &  $F_{[0,15]} G_{[0,10]}(R_1) \land \land G_{[0,30]} (\neg O_1)$ & R+A+SB\\
\hline
 $\phi_3$ &  $F_{[0,15]} (R_1 \land F_{[0,15]} (R_2))$  &SV \\
\hline
\end{tabular}
\vspace{10pt}
\caption{Goal STL specifications. Here, MRC: Multi Reach Choice, A: Avoid, SB: Stabilization, SV: Sequenced Visit. $R_1$, $R_2$, $R_3$ encode the multiple target regions, while $O_1$ encodes the obstacle.}

\label{tab:specs}

\end{table*}

\subsubsection{Tokenization} Our state and action tokenizers are designed to take as input raw states and continuous actions similar to \cite{bonatti2023pact}. For the STL specification, we leverage state-of-the-art text tokenizers like Contrastive Language-Image Pre-Training (CLIP)~\cite{radford2021learning} and Bidirectional Encoder Representations from Transformers (BERT)~\cite{devlin2018bert}. Due to the natural language like syntax of STL, these tokenizers provide richly grounded features that can be leveraged downstream for trajectory predictions. 

Due to STL's rich syntax, there are multiple ways to represent STL specifications as a textual input to the tokenizers. A general technique is to represent the STL specification as an Abstract Syntax Tree and perform different tree traversals (in order, post order, pre order) to generate linear text. Additional, since STL's temporal operators can be represented in multiple formats ("F", "eventually", "finally"), the search space for tokenization techniques is expanded further. Authors in \cite{chen2023nl2tl} did an analysis on different expression formats and converged on in order traversal plus word representation for target outputs. However, in our work the logical specifications are provided as input to PASTEL, so arbitary tokenization procedures can significantly impact the predicted state action trajectory. Additionally, since our focus is on motion planning, atomic propositions cannot be represented using words and require mathematical expressions to capture different regions in the space. Due to these reasons, we choose to finetune SOTA pretrained text tokenizers for our framework. Tokenizer fine-tuning is done via training the entire architecture, including the tokenizers, end-to-end, rather than freezing the tokenizer layers and using their outputs downstream.
\subsection{Model Architecture}

Following the implementation in \cite{bonatti2023pact}, we use a decoder-only Transformer model to roll out a sequence of state-action-specification triplets constituting a trajectory. We leverage a causal attention mask to enforce dependence of future state and action tokens on only past state action and specification tokens. We factor in the specification through two key techniques: First, in order to encode emphasis on the specification while generating trajectories, we repeat the specification token and append it to each state action token at each timestep in a trajectory. We call this technique \textit{specification conditioned prediction}. For this, we use an additional MLP layer to reduce the dimensionality of the feature outputs from STL tokenizer to the embedding space dimension. Second, similar to VLAs we perform a cross attention operation with state, action, specification embeddings acting as Key and Value arrays while the specification embeddings act as Query array. This forces the model to not overfit solely on the state action data and prevents ignoring specification while making predictions.  

PASTEL outputs 3N tokens where N is the time horizon (number of timesteps) of the STL specification, and the prediction involves outputting state, action and specification tokens at each timestep. Since STL specifications encode temporal behavior, the output tokens are made to conform to the different parts of the specification that currently need to be satisfied depending on the timestep. 

\subsection{Implementation Details}

\subsubsection{Training objective}
Our training objective is carefully designed to account for the expectation of high precision state action trajectory predictions. We use a combination of Mean Absolute Error (
MAE) and Mean Squared Error (MSE) loss defined over the state as well as action predictions. Formally, 

\begin{equation}
    \mathcal{L}_{state} = MSE({s}_{t_{obs}},{\hat{s}}_{t_{obs}}) + MAE({s}_{t_{obs}},{\hat{s}}_{t_{obs}})
\end{equation}

\begin{align}
    \mathcal{L}_{action} = MSE({a}_{t},{\hat{a}}_{t}) + MAE({a}_{t},{\hat{a}}_{t}) 
\end{align}

Additionally, we also add a \textit{specification relevance loss} that penalizes deviation from the expected impact of the specification on the predicted trajectory. This loss is defined over the specification embeddings that capture the semantic information about the specification and the cross attention outputs that integrates information from all the input embeddings. Specifically, we use a cosine similarity measure to quantify how well the instruction is reflected in the final outputs. Formally,

\begin{align}
    \mathcal{L}_{spec} =1 -  Cos\_sim({T}_{emb},{C}) 
\end{align}

where \(\mathbf{\textit{$T_{emb}$}}\) represents the mean text embeddings and \(\mathbf{\textit{C}}\) represents the mean cross-attention output embeddings:

\[
\mathbf{T_{emb}} = \frac{1}{n} \sum_{i=1}^n \mathbf{t}_i \quad \text{and} \quad \mathbf{C} = \frac{1}{n} \sum_{i=1}^n \mathbf{c}_i
\]

Here, \(n\) is the number of embeddings (e.g., batch size), \(\mathbf{t}_i\) are the text embeddings, and \(\mathbf{c}_i\) are the cross-attention output embeddings.

Finally, our total loss is:

\begin{equation}
    \mathcal{L}_{total} = \mathcal{L}_{state} + \mathcal{L}_{act} + \mathcal{L}_{spec}
\end{equation}

\subsubsection{Dataset and Model Parameters}
Our dataset is generated by leveraging the Mixed Integer Convex Program (MICP) encodings of STL specifications. We use STLPy~\cite{kurtz2022mixed} to generate a multi specification dataset with varying initial states and complexity of behaviors.  STLPy is a framework that takes as input system dynamics, specification description, actuation constraints and state cost functions to generate trajectories that satisfy the given specifications. For a more formal proof of soundness and other details,  we refer the readers to ~\cite{kurtz2022mixed}.

We outline the specifications in Table 1. Our environment is a 2D planar environment with multiple goal regions and a single obstacle. The atomic propositions are defined over these regions. The specifications over these atomic propositions are designed to encode different robotic mission patterns \cite{DBLP:journals/corr/abs-1901-02077} such as sequenced visit, stabilization etc. These class of patterns capture a large set of common robot mission requirements as identified in \cite{DBLP:journals/corr/abs-1901-02077}.

Our model has similar architecture and parameter configurations as the PACT model. We refer the readers to \cite{bonatti2023pact} for more information. The cross attention operation was added on top of the causal attention operation along with an additional layer of normalization over the outputs.



\section{Evaluation}

\subsection{Experimental Setup}

\begin{table}[h!]
\centering
\begin{tabular}{c|ccc}
\hline
\textbf{Model} & \multicolumn{3}{c}{\textbf{Percentage Satisfaction}} \\
\hline
 & \textbf{$\phi_1$} & \textbf{$\phi_2$} & \textbf{$\phi_3$}  \\
\hline
PACT  & 14 & 41  &  46\\
\hline
PASTEL  & 26  & 75  & 71 \\
\hline
\end{tabular}
\caption{PASTEL satisfaction benchmarking against Vanilla PACT. The values represent the ratio of generated trajectories that completely satisfy the STL specifications out of the the total generated trajectories.}
\label{table:simple}
\end{table}
We evaluate PASTEL\footnote{https://github.com/ScaledFoundations/PACT\_STL}  against the baseline PACT implementation as provided in \cite{bonatti2023pact} . Both the models were trained on the same dataset comprising a total of 20000 state action trajectories across all the specifications. All experiments were run on a workstation with a NVIDIA GeForce RTX 4070 GPU.

The two main research questions we investigate in this paper are:
\begin{enumerate}
    \item \textbf{RQ1}: Does PASTEL achieve higher specification satisfaction compared to Vanilla PACT?
    \item \textbf{RQ2}: Does modifying the text specification at test time have an impact on the final trajectory?
\end{enumerate}

While RQ1 can be measure quantitatively through sampling, RQ2 is evaluated qualitatively by modifying the specification string manually. RQ1 investigates improvement in satisfaction while RQ2 investigates if the model is memorizing the state action trajectories without any dependence on the text specification.

\subsection{Results and Discussion}

Table 2 summarizes the benchmarking results for our experimental setup. The \% satisfaction is computed out of 100 trajectories generated by the model from 100 randomly sampled states. We observe that for $\phi_2$ and $\phi_3$, PASTEL improves satisfaction over baseline by approximately 82.9 and 54.3 percent respectively. For $\phi_1$, we observe an 85.7 percent performance improvement over baseline. Additionally, all the generated trajectories are smooth and respect the actuation constraints measured separately via examining the generated actions.

As can be inferred, PASTEL has medium to high satisfaction for $\phi_2$ and $\phi_3$  while the PACT has low to medium satisfaction. We hypothesize this is due to the relative complexity of $\phi_2$ and $\phi_3$ involving two nested tasks (reaching and staying in a region and reaching two regions) respectively  which is complex for autoregressive models like PACT to infer without additional contextual information. Due to our conditional prediction and cross attention mechanism, the model is able to leverage the specification embeddings to satisfy the task succesfully. However, both PASTEL and PACT's satisfaction percentage drops for $\phi_1$ which is the most complex specification involving a disjunction operator. This implies that  there are multiple possible ways to satisfy the requirement and the underlying dataset reflects the same. Nonetheless, PASTEL still outperforms PACT for $\phi_1$ relatively. Hence, we definitively answer RQ1. Additionally, the generated trajectories satisfy the actuation constraints which was never explicitly encoded into the model design. This implicit effect is due to the loss function defined over the ground truth actions that respect the actuation constraints due to the optimisation problem setup. This lends additional support to the power of autoregressive trajectory generators in generating safe constraint satisfying trajectories.

For RQ2, we modified specification text and visualised attention matrix to observe dependence on text. We underline our two key qualitative insights:

\begin{enumerate}
    \item The satisfaction drops when the test specification is different than the training one in terms of the atomic propositions encoding the regions.
    \item The attention matrix visualisation highlights the dependence of state and action tokens on the specification tokens further substantiating our original hypothesis.
\end{enumerate}




\section{Related Work}
Constrained trajectory generation for robots has been a focal point in robotics research, particularly in applications requiring high precision and adherence to strict operational constraints. Traditional methods have relied on optimization-based approaches, such as Mixed-Integer Linear Programming (MILP) and Sequential Quadratic Programming (SQP), which are effective but often computationally intensive and challenging to scale for complex tasks. Sampling-based planners like Rapidly-exploring Random Trees (RRT)~\cite{LaValle1998RapidlyexploringRT, karaman2011sampling} and Probabilistic Roadmaps (PRM)~\cite{508439} have been used to address these issues from primarily a collision-avoidance perspective, introducing probabilistic guarantees for constraint satisfaction. Constrained variants of RRT-like methods have also been proposed in the literature~\cite{5152399}. However, these methods can struggle with high-dimensional spaces and intricate constraints, limiting their applicability in real-world scenarios. Another class of techniques involves the usage of control barrier functions: learning safe policies using explicit control barrier functions~\cite{chen2017obstacle, patrikar2022challenges}, or constructing CBFs jointly with the policy using neural networks~\cite{meng2021reactive, dawson2022learning}.

Data-driven trajectory planning techniques mainly adapt two paradigms: reinforcement learning and imitation learning. While reinforcement learning has been widely used to learn safe policies, the success of these methods often depends on manual reward shaping, which is a laborious and non-trivial effort, as well as the existence of a capable simulator or sandbox that allows for a large number of training episodes. Imitation learning has the potential to reduce sample complexity using a more focused set of demonstrations, for example, in the case of safe trajectory planning, learning only from a set of safe trajectories. Imitation learning can take the form of simple Behavior Cloning~\cite{bain1995framework}, which may not generalize to the out-of-distribution scenarios induced by
on-policy deployment, and often requires additional training~\cite{ross2011reduction}. Other methods such as inverse reinforcement learning~\cite{ng2000algorithms} can mitigate this challenge but they do not explicitly account for constraints either and rely on implicit extraction of a reward signal apparent in the data. 

In recent work, reinforcement learning and imitation learning have been posed as sequence modeling problems to leverage the immense efficacy of Transformer models at learning from data. One seminal work was the Decision Transformer~\cite{chen2021decision}, which uses a causally masked Transformer conditioned on desired rewards to output optimal trajectories. Similarly, works such as Gato, PACT use Transformer models directly on demonstrations to learn trajectories. Extensions such as ConBAT ~\cite{meng2023conbat} attempt safe trajectory planning by training on a combination of safe and unsafe demonstrations. 

\section{Conclusion and Future Work} 
\label{sec:conclusion}

In this work, we propose a novel approach to factor in STL specifications in transformer based trajectory planning using specification conditioned prediction and cross attention based mechanism for specification relevance. 

While, our initial results are promising our approach currently suffers with specifications with long horizons and complex nested tasks. We plan to remedy this using the STL decomposition techniques proposed in \cite{kapoor2024safe} and updating the specification token at each timestep to only focus on the relevant part of the specification. Additionally, we also plan to factor in external feedback in the form of STL robustness similar to SMART\cite{sun2023smart}. Finally, we aim to evaluate the feasibility of our technique via field testing on navigation robots.

\section*{Acknowledgments}


\bibliographystyle{unsrt}
\bibliography{references}

\begin{thebibliography}{10}

\bibitem{clark2016faulty}
Jack Clark and Dario Amodei.
\newblock Faulty reward functions in the wild.
\newblock {\em OpenAI Codex}, 2016.

\bibitem{vemprala2024chatgpt}
Sai~H Vemprala, Rogerio Bonatti, Arthur Bucker, and Ashish Kapoor.
\newblock Chatgpt for robotics: Design principles and model abilities.
\newblock {\em IEEE Access}, 2024.

\bibitem{tellex2020robots}
Stefanie Tellex, Nakul Gopalan, Hadas Kress-Gazit, and Cynthia Matuszek.
\newblock Robots that use language.
\newblock {\em Annual Review of Control, Robotics, and Autonomous Systems}, 3:25--55, 2020.

\bibitem{Maler2004MonitoringTP}
Oded Maler and D.~Nickovic.
\newblock Monitoring temporal properties of continuous signals.
\newblock In {\em FORMATS/FTRTFT}, 2004.

\bibitem{aksaray2016qlearning}
Derya Aksaray, Austin Jones, Zhaodan Kong, Mac Schwager, and Calin Belta.
\newblock Q-learning for robust satisfaction of signal temporal logic specifications, 2016.

\bibitem{kapoor2020model}
Parv Kapoor, Anand Balakrishnan, and Jyotirmoy~V Deshmukh.
\newblock Model-based reinforcement learning from signal temporal logic specifications.
\newblock {\em arXiv preprint arXiv:2011.04950}, 2020.

\bibitem{kurtz2022mixed}
Vince Kurtz and Hai Lin.
\newblock Mixed-integer programming for signal temporal logic with fewer binary variables.
\newblock {\em arXiv preprint arXiv:2204.06367}, 2022.

\bibitem{kapoor2024safe}
Parv Kapoor, Eunsuk Kang, and R{\^o}mulo Meira-G{\'o}es.
\newblock Safe planning through incremental decomposition of signal temporal logic specifications.
\newblock In {\em NASA Formal Methods Symposium}, pages 377--396. Springer, 2024.

\bibitem{10160953}
Jasmine~Jerry Aloor, Jay Patrikar, Parv Kapoor, Jean Oh, and Sebastian Scherer.
\newblock Follow the rules: Online signal temporal logic tree search for guided imitation learning in stochastic domains.
\newblock In {\em 2023 IEEE International Conference on Robotics and Automation (ICRA)}, pages 1320--1326, 2023.

\bibitem{leungbackpropagation}
Karen Leung, Nikos Ar{\'e}chiga, and Marco Pavone.
\newblock Backpropagation through signal temporal logic specifications: Infusing logical structure into gradient-based methods.
\newblock {\em The International Journal of Robotics Research}, page 02783649221082115, 2023.

\bibitem{bonatti2023pact}
Rogerio Bonatti, Sai Vemprala, Shuang Ma, Felipe Frujeri, Shuhang Chen, and Ashish Kapoor.
\newblock Pact: Perception-action causal transformer for autoregressive robotics pre-training.
\newblock In {\em 2023 IEEE/RSJ International Conference on Intelligent Robots and Systems (IROS)}, pages 3621--3627. IEEE, 2023.

\bibitem{ma2024survey}
Yueen Ma, Zixing Song, Yuzheng Zhuang, Jianye Hao, and Irwin King.
\newblock A survey on vision-language-action models for embodied ai.
\newblock {\em arXiv preprint arXiv:2405.14093}, 2024.

\bibitem{radford2021learning}
Alec Radford, Jong~Wook Kim, Chris Hallacy, Aditya Ramesh, Gabriel Goh, Sandhini Agarwal, Girish Sastry, Amanda Askell, Pamela Mishkin, Jack Clark, et~al.
\newblock Learning transferable visual models from natural language supervision.
\newblock In {\em International conference on machine learning}, pages 8748--8763. PMLR, 2021.

\bibitem{devlin2018bert}
Jacob Devlin, Ming-Wei Chang, Kenton Lee, and Kristina Toutanova.
\newblock Bert: Pre-training of deep bidirectional transformers for language understanding.
\newblock {\em arXiv preprint arXiv:1810.04805}, 2018.

\bibitem{chen2023nl2tl}
Yongchao Chen, Rujul Gandhi, Yang Zhang, and Chuchu Fan.
\newblock Nl2tl: Transforming natural languages to temporal logics using large language models.
\newblock {\em arXiv preprint arXiv:2305.07766}, 2023.

\bibitem{DBLP:journals/corr/abs-1901-02077}
Claudio Menghi, Christos Tsigkanos, Patrizio Pelliccione, Carlo Ghezzi, and Thorsten Berger.
\newblock Specification patterns for robotic missions.
\newblock {\em CoRR}, abs/1901.02077, 2019.

\bibitem{LaValle1998RapidlyexploringRT}
Steven~M. LaValle.
\newblock Rapidly-exploring random trees : a new tool for path planning.
\newblock {\em The annual research report}, 1998.

\bibitem{karaman2011sampling}
Sertac Karaman and Emilio Frazzoli.
\newblock Sampling-based algorithms for optimal motion planning.
\newblock {\em The international journal of robotics research}, 30(7):846--894, 2011.

\bibitem{508439}
L.E. Kavraki, P.~Svestka, J.-C. Latombe, and M.H. Overmars.
\newblock Probabilistic roadmaps for path planning in high-dimensional configuration spaces.
\newblock {\em IEEE Transactions on Robotics and Automation}, 12(4):566--580, 1996.

\bibitem{5152399}
Dmitry Berenson, Siddhartha~S. Srinivasa, Dave Ferguson, and James~J. Kuffner.
\newblock Manipulation planning on constraint manifolds.
\newblock In {\em 2009 IEEE International Conference on Robotics and Automation}, pages 625--632, 2009.

\bibitem{chen2017obstacle}
Yuxiao Chen, Huei Peng, and Jessy Grizzle.
\newblock Obstacle avoidance for low-speed autonomous vehicles with barrier function.
\newblock {\em IEEE Transactions on Control Systems Technology}, 26(1):194--206, 2017.

\bibitem{patrikar2022challenges}
Jay Patrikar, Joao Dantas, Sourish Ghosh, Parv Kapoor, Ian Higgins, Jasmine~J Aloor, Ingrid Navarro, Jimin Sun, Ben Stoler, Milad Hamidi, et~al.
\newblock Challenges in close-proximity safe and seamless operation of manned and unmanned aircraft in shared airspace.
\newblock {\em arXiv preprint arXiv:2211.06932}, 2022.

\bibitem{meng2021reactive}
Yue Meng, Zengyi Qin, and Chuchu Fan.
\newblock Reactive and safe road user simulations using neural barrier certificates.
\newblock In {\em 2021 IEEE/RSJ International Conference on Intelligent Robots and Systems (IROS)}, pages 6299--6306. IEEE, 2021.

\bibitem{dawson2022learning}
Charles Dawson, Bethany Lowenkamp, Dylan Goff, and Chuchu Fan.
\newblock Learning safe, generalizable perception-based hybrid control with certificates.
\newblock {\em IEEE Robotics and Automation Letters}, 7(2):1904--1911, 2022.

\bibitem{bain1995framework}
Michael Bain and Claude Sammut.
\newblock A framework for behavioural cloning.
\newblock In {\em Machine Intelligence 15}, pages 103--129, 1995.

\bibitem{ross2011reduction}
St{\'e}phane Ross, Geoffrey Gordon, and Drew Bagnell.
\newblock A reduction of imitation learning and structured prediction to no-regret online learning.
\newblock In {\em Proceedings of the fourteenth international conference on artificial intelligence and statistics}, pages 627--635. JMLR Workshop and Conference Proceedings, 2011.

\bibitem{ng2000algorithms}
Andrew~Y Ng, Stuart Russell, et~al.
\newblock Algorithms for inverse reinforcement learning.
\newblock In {\em Icml}, volume~1, page~2, 2000.

\bibitem{chen2021decision}
Lili Chen, Kevin Lu, Aravind Rajeswaran, Kimin Lee, Aditya Grover, Misha Laskin, Pieter Abbeel, Aravind Srinivas, and Igor Mordatch.
\newblock Decision transformer: Reinforcement learning via sequence modeling.
\newblock {\em Advances in neural information processing systems}, 34:15084--15097, 2021.

\bibitem{meng2023conbat}
Yue Meng, Sai Vemprala, Rogerio Bonatti, Chuchu Fan, and Ashish Kapoor.
\newblock Conbat: Control barrier transformer for safe policy learning.
\newblock {\em arXiv preprint arXiv:2303.04212}, 2023.

\bibitem{sun2023smart}
Yanchao Sun, Shuang Ma, Ratnesh Madaan, Rogerio Bonatti, Furong Huang, and Ashish Kapoor.
\newblock Smart: Self-supervised multi-task pretraining with control transformers.
\newblock {\em arXiv preprint arXiv:2301.09816}, 2023.

\end{thebibliography}

\end{document}